\documentclass[10pt, a4paper]{article}

\usepackage{lrec-coling2024} 
\usepackage{fancyhdr}

\usepackage[colorinlistoftodos]{todonotes}
\usepackage{booktabs}
\usepackage{tikz}
\usepackage{graphicx}
\usepackage{multirow}
\usepackage{hyperref}
\usepackage{tabularx}
\usepackage{orcidlink}
\usepackage{marvosym}
\newcolumntype{R}{>{\raggedleft\arraybackslash}X}

\usetikzlibrary{shapes,arrows,positioning}
\usepackage{enumitem}

\newcommand{\ares}[2]{$#1 \pm #2$}

\newcommand{\pos}[1]{{\color{purple} #1}}
\newcommand{\np}[1]{{\color{teal} #1}}
\newenvironment{myquote}%
  {\list{}{\leftmargin=0.1in\rightmargin=0.1in}\item[]}%
  {\endlist}

\newcommand{\FRorcid}{\orcidlink{0000-0002-1697-8586}}
\newcommand{\AMorcid}{\orcidlink{0000-0002-3387-6557}}
\newcommand{\CTorcid}{\orcidlink{0000-0001-6976-3258}}
\newcommand{\ABCorcid}{\orcidlink{0000-0003-4719-3420}}

\title{PejorativITy: Disambiguating Pejorative Epithets \\ to Improve Misogyny Detection in Italian Tweets\\}

\name{Arianna Muti$^1$\textsuperscript{\Letter}\AMorcid,
Federico Ruggeri$^2$\FRorcid,
Cagri Toraman$^3$\CTorcid, \\
\bf \large Lorenzo Musetti$^4$,
\bf \large Samuel Algherini$^4$,
\bf \large Silvia Ronchi$^4$, \\
\bf \large Gianmarco Saretto$^4$, 
\bf \large Caterina Zapparoli$^4$,
\bf \large Alberto Barrón-Cedeño$^1$\ABCorcid}

\address{$^1$DIT, University of Bologna, Forlì, Italy \\
    $^2$DISI, University of Bologna, Bologna, Italy \\
    $^3$CEng, Middle East Technical University, Ankara, Turkey \\
    $^4$Expert AI, Modena, Italy \\
    \texttt{\{arianna.muti2, federico.ruggeri6, a.barron\}@unibo.it} \\
    \texttt{ctoraman@ceng.metu.edu.tr} \\
    \texttt{info@samuelalgherini.com}\\
    \texttt{\{lmusetti, sronchi, czapparoli, gsaretto\}@expert.ai}}

\abstract{%
Misogyny is often expressed through figurative language. Some neutral words can assume a negative connotation when functioning as pejorative epithets. Disambiguating the meaning of such terms might help the detection of misogyny. In order to address such task, we present PejorativITy, a novel corpus of 1,200 manually annotated Italian tweets for pejorative language at the word level and misogyny at the sentence level.
We evaluate the impact of injecting information about disambiguated words into a model targeting misogyny detection.
In particular, we explore two different approaches for injection: concatenation of pejorative information and substitution of ambiguous words with univocal terms.
Our experimental results, both on our corpus and on two popular benchmarks on Italian tweets, show that both approaches lead to a major classification improvement, indicating that word sense disambiguation is a promising preliminary step for misogyny detection.
Furthermore, we investigate LLMs' understanding of pejorative epithets by means of contextual word embeddings analysis and prompting.
\\ \newline 
}

\begin{document}

\maketitleabstract
 \noindent \textbf{Disclaimer}: This paper contains examples of offensive and explicit content.

\thispagestyle{fancy} 

\section{Introduction}

Pejorative language refers to a word or phrase that has negative connotations and is intended to disparage or belittle.\footnote{\url{https://www.merriam-webster.com/dictionary/pejorative}} An inoffensive word becoming pejorative is a form of semantic drift known as pejoration; thus, pejorativity is context-dependent:
pejorative words have one primary neutral meaning, and another negatively connotated meaning. The opposite is known as melioration, which is when a term begins as pejorative and eventually is adopted in a neutral sense, like in the case of slur reappropriation~\cite{415ef38c-b0ae-3d7a-b56e-3c1428a02804}.
Pejorative words are relevant in misogyny detection since many neutral words are used to address women in an offensive way, targeting either their physical aspect or their intelligence.
We refer to such terms as \textbf{pejorative epithets}.
Some examples in Italian are \textit{balena} (whale/fat woman) and \textit{gallina} (chicken/stupid).
State-of-the-art models struggle to correctly classify misogyny when sentences contain such terms~\cite{Fersini2020}. The occurrence of polysemic words with a pejorative connotation in the training set and a neutral connotation in the test set results in a great number of false positives~\cite{Muti2020UniBOA}.
For this reason, we introduce pejorative epithets disambiguation as a preliminary step to detect misogyny.
Our goal is to assess whether the disambiguation of potentially pejorative epithets improves the detection of misogynistic language, while reducing the rate of false positives.

\begin{figure}[!t]
    \centering
    \includegraphics[width=\columnwidth]{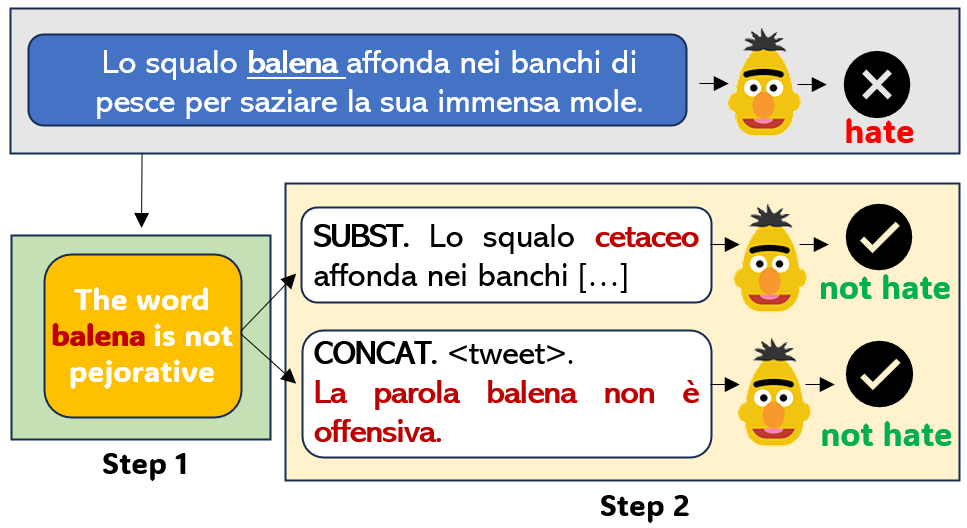}
    \caption{Our pipeline. Step 1: a model identifies the connotation of possibly pejorative epithets. Step 2: the identified connotation is used to enrich (CONCAT) and substitute (SUBST) part of the textual input for misogyny detection.}
    \label{fig:diagramma}
\end{figure}

In this work, we aim to answer three research questions:
\begin{description}
\item[RQ1] Which epithets are used in misogynistic language in Italian?
\item[RQ2] Can the disambiguation of such words decrease the error rate in misogyny detection?
\item[RQ3] Can encoder-based language models and generative LLMs differentiate if a word in a tweet is pejorative or neutral based on its context?
\end{description}

To address \textbf{RQ1}, we compile a list of pejorative words used online to address women. We use such words to 
retrieve new tweets, and build PejorativITy, a novel corpus of Italian tweets, annotated at the word level for pejorativity, and at the sentence level for misogyny.

To address \textbf{RQ2}, we fine-tune two BERT-based models: model$_{pej}$ to identify whether a word in the context of a tweet is pejorative or neutral, following~\citet{dinu-etal-2021-computational-exploration}, and model$_{mis}$ to detect misogyny.
We use the output of model$_{pej}$ to inform model$_{mis}$ of whether the target word is pejorative within that context or not.
Figure \ref{fig:diagramma} represents our pipeline. 

To address \textbf{RQ3}, we compare the cosine similarity between the contextualised word embeddings of a BERT-based model and their univocal corresponding words (\textbf{anchors}) before and after fine-tuning for pejorativity detection. 
Additionally, we prompt popular instruction-tuned LLMs to test their ability to disambiguate potentially pejorative words based on the context.

Our contribution is threefold: 
(1)~we release a corpus manually annotated for pejorativity at the word level and misogyny at the sentence level; 
(2)~we develop a transformer-based model for detecting pejorative words, whose predictions are used to enhance the performance of the model targeting misogyny detection; and 
(3)~we analyse the performance of SOTA generative models on pejorative epithets disambiguation.

To the best of our knowledge, this is the first work that proposes word sense disambiguation to linguistically inform computational models for misogyny detection.

The PejorativITy dataset and the code for all the experiments
are available at \url{https://github.com/arimuti/PejorativITy}.


\section{Related Work}
\label{sec:append-how-prod}
Misogyny and sexism detection have been explored in different platforms, such as Gab and Reddit~\cite{kirk-etal-2023-semeval,guest-etal-2021-expert}, Twitter~\cite{jha-mamidi-2017-compliment,anzovino2018automatic}, and blogs~\cite{breitfeller_finding_2019} in English; and in different languages, such as Spanish~\cite{anzovino2018automatic,plaza2023exist1}, Arabic~\cite{almanea-poesio-2022-armis}, and Turkish~\cite{toraman2022large}.
In Italian, the reference datasets for the identification of misogyny are the two compiled in the framework of the two editions of the Automatic Misogyny Identification shared task (AMI)~\citelanguageresource{fersininozzarosso-overview-2018,Fersini2020}.

Our work takes inspiration from~\citet{dinu-etal-2021-computational-exploration}, who
(a)~explore pejorative language on social media for the first time;
(b)~build a multilingual lexicon of pejorative terms for English, Spanish, Italian, and Romanian;
(c)~release a dataset of tweets annotated for pejorative use; and
(d)~present an attempt to automatically disambiguate pejorative words in their dataset. Our contribution differs since, for the first time, the information about the pejorativity of a word is leveraged to inform the model for misogyny detection. Moreover, our pejorative lexicon contains words that are currently used on Twitter to address women in a misogynistic manner. 
Whereas~\citeauthor{dinu-etal-2021-computational-exploration}'s lexicon
considers hate speech in general, most gender-based words are outdated or missing, and it does not focus on the sort of slang typically used online.  

Another similar work is~\citet{pamungkas_investigating_2023}, who develop the Swear Words Abusiveness Dataset (SWAD), where abusive swearing in English tweets is manually annotated at the word level to address the task of predicting the abusiveness of a swear word based on its context.
While their work focuses on spotting slurs when used in a neutral way (i.e.\ meliorations), our aim is to disambiguate neutral words used in an offensive way (i.e.\ pejoration).
Moreover, \citeauthor{pamungkas_investigating_2023} exclude highly ambiguous words when creating their target word lexicon, whereas we precisely focus on them.

\begin{table*}[!t]
    \centering
    \small
    \resizebox{1.0\linewidth}{!}{
    \begin{tabular}{lllll}
        \textbf{Word} & \textbf{Literal} & \textbf{Pejorative} & \textbf{Neutral anchor} & \textbf{Pejorative anchor}\\

        \midrule
        \textbf{acida} & acid/sour & peevish & aspra & intrattabile, stronza \\
        \textbf{asina} & female donkey & stupid & ciuco & stupida \\
        \textbf{balena} & whale/flash & fat woman & cetaceo, balenare & grassa\\
        \textbf{bambola} & doll & girl (objectifying) & giocattolo & donna attraente\\
        \textbf{cagna} & female dog & bitch & cane femmina, canide & donna di facili costumi, troia \\
         \textbf{cavalla} & female horse & ugly/whore & equino & brutta, alta e grossa \\
         \textbf{civetta}& owl & tease & volatile rapace & donna che cerca attenzioni\\
        \textbf{cesso}& toilet & ugly & water, bagno, toilette & brutta \\
        \textbf{contadina }& farmer & ignorant, illiterate & agricoltore femmina & donna ignorante \\
        \textbf{cortigiana }& court lady & prostitute & dama di corte & prostituta\\
        \textbf{cozza }& mussel & ugly/clingy & mollusco & donna brutta, appiccicosa \\
         \textbf{femminista} & feminist & feminazi & femminista & polemica, fastidiosa \\
         \textbf{fogna} & sewer & skanky & fognatura & schifosa, bocca\\
         \textbf{gallina} & chicken & stupid & pennuto & stupida\\
         \textbf{grezza} & raw & rude woman & non lavorato & rozza\\
         \textbf{lesbica} & lesbian & lesbian (offensive) & donna a cui piacciono le donne & schifosa \\
         \textbf{lurida} & dirty & skanky & sporca & promiscua, troia \\
          \textbf{maiala} & sow & whore & maiale femmina & promiscua, troia  \\
        \textbf{mucca} & cow & bitch & bovide & stupida, troia \\
         \textbf{oca} & goose & stupid girl & pennuto & stupida, pettegola\\
         \textbf{pecora} & sheep & doormat & ovino & stupida \\
        \textbf{strega}& witch & hag, unpleasant & maga & crudele  \\
        \textbf{vacca} & cow & whore & bovino & donna di facili costumi, troia \\
        \textbf{zingara} & gipsy & shabby & gitana & trasandata \\

        \bottomrule

    \end{tabular}
    }
    \caption{Italian pejorative lexicon, their literal and pejorative translations in English, and their anchors.}
    \label{tab:pejorative_keywords}
\end{table*}

\section{Corpus Compilation} \label{sec:corpus_compilation}

To provide an overview of which misogynous epithets are commonly used on Twitter in Italian (\textbf{RQ1}), we compile a novel corpus.
The compilation 
involves two steps: the creation 
of a lexicon of polysemic words that can function as pejorative epithets for women, and the retrieval of tweets containing such words.

\paragraph{Lexicon.}
We collect our lexicon by selecting words from three distinct sources.
(1) We ask ten Italian native speakers to provide a list of offensive words used online to address women. The speakers use social media on a daily basis and their age ranges between 27 and 39 years.
(2) We retrieve the keywords used in the two Italian corpora for the Automatic Misogyny Identification (AMI) shared task~\cite{fersininozzarosso-overview-2018, Fersini2020}.
3) We consult the 'List of Dirty Naughty Obscene Bad Words'.\footnote{\href{https://github.com/LDNOOBW/List-of-Dirty-Naughty-Obscene-and-Otherwise-Bad-Words/tree/master}{https://github.com/LDNOOBW/List-of-Dirty-Naughty-Obscene-and-Otherwise-Bad-Words/tree/master}, consulted on January 2023.}
We only keep polysemic words whose primary meaning is neutral and that are frequently used on Twitter with both pejorative and neutral connotations.
To ensure the quality of our vocabulary, we qualitatively verify that such words are used with both connotations by manually searching them on Twitter.\footnote{Due to their exclusive neutral or negative connotation on Twitter,
the following words are discarded: \textit{barile, banco, botte, barbona, facile, gatta morta, passeggiatrice, porca, principessa, privilegiata, psicopatica, scrofa, somara, travestita}.}

Table~\ref{tab:pejorative_keywords} shows our lexicon of 24 words.
For each word, we report the English translation of its literal and pejorative meaning, and their anchors in Italian.
Anchor words refer to the unambiguous words used to define polysemic words.
We call these words anchors because their meaning is univocal and does not change according to the context. 
For instance, the word \textit{balena (whale)} is used to refer to either a sea mammal or an overweight woman.
In contrast, the anchor words \textit{cetaceo (cetacean)} and \textit{grassa (fat)} only refer to the animal in the first case and to being overweight in the second case, at least as far as their use in Twitter is concerned.\footnote{In this case, the word \textit{balena} has a third anchor word, from the verb \textit{balenare}, which means 'to flash'.}

\paragraph{Tweets.}
We use Twarc\footnote{\url{https://twarc-project.readthedocs.io}} to retrieve tweets from December 2022 to February 2023 containing words in our lexicon.
We select 50 tweets for each word in our lexicon, resulting in 1,200 tweets.
We keep a balance of pejorative and neutral use of lexicon words, although an equal distribution for each word could not be guaranteed.

\begin{table*}[!t]

\centering
\begin{tabular}{lp{0.8\columnwidth}p{0.95\columnwidth}}
\toprule
\bf ID & \bf Tweet & \bf Translation \\
\midrule
\textit{70019}  & \textit{Non voglio una \underline{cagna} un cane ce l'ho giaaaa} & \textit{I don't want a \underline{female dog/bitch}, I have a dog already.} \\[0.1cm] 

\textit{30021}  & \textit{Wow sei una \underline{bambola}!}  & \textit{Wow you're a \underline{beautiful girl/doll!}}\\[0.1cm] 

\textit{10010}  & \textit{Xchè avrà dato una risposta  \underline{acida} a lui}
 & \textit{Because he/she will have given him a \underline{sharp} answer} \\[0.1cm] 
 
\textit{61209}  & \textit{Ma come fai a dire che sei una \underline{balena} sei bellissima} & \textit{How can you say you're \underline{a whale/fat}, you're beautiful} \\
\bottomrule
\end{tabular}
\caption{Examples of tweets with potentially pejorative words (\underline{underlined}).}
\label{tab:examples}
\end{table*}

\section{Data Annotation}

We recruit six annotators with a background in linguistics, gender studies, cognitive sciences, and NLP to label our corpus for pejorative word disambiguation and misogyny detection.

We first devise a pilot annotation study to explore the complexity of the task and observe differences in how male and female annotators perceive pejorative connotations.
For this purpose, we follow a descriptive annotation paradigm~\cite{RottgerVHP22}, which encourages annotator subjectivity by not providing guidelines.
We split the annotators into two groups and assign 50 tweets each for labeling.
Each group is composed of two women and one man with ages ranging between 27 and 39 years old.

We use Krippendorff's alpha~\cite{Krippendorff2011Computing} to measure the inter-annotator agreement (IAA).
The IAA of the first group is \textit{moderate} for both pejorativity (0.48) and misogyny (0.50), whereas the IAA of the second group is \textit{fair} for pejorativity (0.33) and \textit{moderate} for misogyny (0.50).
We observe that, in terms of gender differences, men tend to consider sexual objectifying compliments as non-pejorative.
Based on annotators' feedback, we identify five major areas of disagreement:

\paragraph{Lack of context.} Some tweets are very short, lacking enough context to understand the intention of the author. We decide to label such tweets as neutral.
Consider tweet \textit{70019} in Table~\ref{tab:examples}.

Although it is likely that the author uses humour to address a woman as a \textit{cagna} (bitch), the context does not allow for a clear interpretation: it is possible that the author does not want another (female) dog, because he has already one.

\paragraph{Objectifying compliments.} Some tweets are intended to compliment women, by means of objectification. Thus, we label them as pejorative.
In the tweet \textit{30021} in Table~\ref{tab:examples}, the term \textit{bambola} is used as a compliment, but it is objectifying and, therefore, should be considered pejorative.

\paragraph{Pejorative epithets towards objects.} Some words are used pejoratively towards inanimate objects, therefore, they should be labeled as neutral. 
In the tweet \textit{10010} in Table~\ref{tab:examples}, the term \textit{acida} refers to an inanimate thing (an answer), although the term is used pejoratively.

\paragraph{Pejorative epithets towards men.} Words that are used pejoratively against men should be labeled as pejorative, so that the corpus can be used for the general task of pejorativity detection regardless of the auxiliary task.

\paragraph{Reported Speech.} Some tweets contain pejorative epithets, although the intention is not harmful, because they are contained in reported speech. We label them as pejorative, since the annotation refers to the word, not to the whole sentence.
Consider tweet \textit{61209} in Table~\ref{tab:examples}: the word \textit{balena} is pejorative, but it is used in a positive way by means of negation.
\medskip

We devise a second pilot annotation, getting closer to a prescriptive annotation paradigm~\cite{RottgerVHP22}, by providing the above guidelines to the annotators.
We select the top 50 tweets that caused more debate during the first study annotation phase.
The IAA computed on all six annotators is 0.53 (\textit{moderate}), denoting an improvement over the first pilot study.

\paragraph{PejorativITy.} After the pilot studies, we annotate our collected corpus of 1,200 tweets. Only one person carries out the whole annotation process. We select the annotator with the most interdisciplinary background, who is an expert in gender studies, linguistics and NLP, who has been a target of misogyny.
This setting is considered among the best practices for the annotation of phenomena like pejorative epithets and misogyny~\cite{abercrombie-etal-2023-resources}.

Table~\ref{tab:stats} shows the statistics of our corpus.
The Pearson correlation between misogyny and pejorativity labels is 0.70, which is in line with our expectations. The tweets for which misogyny and pejorativity are not aligned are mainly reported speech or men-related offensive language.
It is worth noting that some sentences might not be considered misogynous, as they do not express hate towards women.
However, they might be considered sexist.
For instance, the sentence \textit{``che bella bambola ciao tesoro''}\footnote{translation: what a beautiful doll (girl), hi darling} does not express hate but perpetuates the objectification of women by addressing the target of the tweet as a doll, falling into the category of benevolent sexism~\cite{gothreau_hostile_2022}.

\begin{table}
\begin{tabularx}{\columnwidth}{lccc}
\toprule
\bf Class       & \bf Training & \bf Test & \bf Total \\
\midrule
Misogynous          & 369     & 28  & 397\\
~~~ Pejorative      & 363 & 28 & 391\\
~~~ Not pejorative  & \,\,\,\,\,6 & -- & \,\,\,\,\,6\\
\midrule
Non-misogynous      & 735   & 68    & 803\\
~~~ Pejorative      & 172 & 18 & 190\\
~~~ Not pejorative  & 563 & 50 & 613\\

\bottomrule
\end{tabularx}
    \caption{Statistics of the PejorativITy corpus. The same tweets are annotated for misogyny and pejorativity, for a total of 1,200 instances. For both the misogynous and the non-misogynous tweets, we report how many contain a pejorative word and how many do not.}
  \label{tab:stats}
\end{table}




\section{Experiments}

To understand the impact of disambiguating pejorative words for misogyny detection (\textbf{RQ2}), we experiment with AlBERTo~\cite{PolignanoEtAlCLIC2019}, a popular BERT-based model trained on $200\,M$ Italian tweets.
In particular, we fine-tune AlBERTo on two downstream tasks: pejorative word disambiguation and misogyny detection.

For pejorativite word disambiguation, we evaluate AlBERTo only on our corpus.
For misogyny detection, we also consider the two other benchmark datasets for Italian: AMI-2018~\cite{fersininozzarosso-overview-2018} and AMI-2020~\cite{Fersini2020}.
To the best of our knowledge, these are the only corpora that address misogyny detection on Italian tweets.
Table~\ref{tab:corpus} shows their statistics.

\begin{table}[t]
\centering

\begin{tabularx}{\columnwidth}{lRRR}
\toprule

\bf AMI-2018 & \bf Misogynous & \bf Not  & \bf Total\\
\midrule
Train    & 1,828   & 2,172    & 4,000 \\
Test      & 512    & 488   & 1,000 \\ 
\midrule
\bf AMI-2020 & \bf Misogynous & \bf Not  & \bf Total\\
\midrule
Train    & 2,337    & 2,663  & 5,000 \\
Test      & 500    & 500  & 1,000 \\
\bottomrule
\end{tabularx}

\caption{Statistics of the AMI 2018 and 2020 corpora~\cite{fersininozzarosso-overview-2018,Fersini2020}.}
\label{tab:corpus}
\end{table}

We formulate the disambiguation of pejorative words as a binary classification task, where a model classifies a word contained in a sentence as pejorative or neutral. Then, we use the information about the pejorativity of a word to enrich the input to the model responsible for the detection of misogyny.
Since AMI-2018 and AMI-2020 are not annotated for pejorative word disambiguation, we use the model fine-tuned on our corpus to determine the connotation of ambiguous words.

Formally, we devise the following pipeline, where $w\in W$ is a word from our lexicon $W$ of pejorative words:
\begin{enumerate}[label=(\alph*)]
\item We train model$_{pej}$ that, given a tweet containing a word $w\in W$, predicts whether $w$ is being used in a pejorative way.
\item We enrich input tweets in all data partitions by injecting knowledge about the pejorativity of our lexicon words according to model$_{pej}$. We try two different approaches to modify the input data: \textit{i)} we \textbf{concatenate} the information about the pejorativity of $w$  
at the end of the tweet or \textit{ii)} we \textbf{substitute} the ambiguous $w$ 
with its  
corresponding anchor word.   

\item We train model$_{mis}$ to detect misogyny with the enriched input tweets.
\end{enumerate}

Our pipeline is meant to process any tweet. However, as a first step, we check whether it contains at least one $w\in W$.    
In our setup, when testing on the subset of AMI-2018 and AMI-2020 containing only pejorative words (epithets), thhat are recognized through string matching after lemmatization.

As hyper-parameters, we use the AdamW optimizer with $\epsilon=1^{-8}$~\cite{2017arXiv171105101L}. We fine-tune AlBERTo for 4 epochs with batch size 16. 
We report macro and per-class F$_1$-measure as standard metrics for binary classification tasks, averaged over three individual runs.
All the experiments are run using Google Colab's GPU.

\begin{table}[!t]

    \centering
\begin{tabularx}{\columnwidth}{lRRR}
\toprule
\textbf{Approach} & \bf Macro & \bf Mis.  & \bf Not \\
\midrule

baseline & 0.68 & 0.56 & 0.79 \\ \midrule
concatenation \\
~~~ w/ gold & 0.83 &0.78 & 0.88 \\
~~~ w/ predictions & 0.75 &0.68 & 0.82 \\ \midrule
substitution  \\
~~~ w/ gold &\bf 0.87 & \bf 0.82 & \bf 0.92 \\
~~~ w/ predictions & 0.77 & 0.69 & 0.84 \\
         \bottomrule
    \end{tabularx}
     \caption{Macro and per-class F$_1$-score on PejorativITy concerning misogyny detection.}
    \label{tab:results_pej}
\end{table}

\section{Results}

Regarding pejorative word disambiguation, the fine-tuned AlBERTo model (model$_{pej}$) reaches a macro F$_1$-measure of $0.82 \pm 0.03$ on the PejorativITy test partition.


Table~\ref{tab:results_pej} shows the classification performance for misogyny detection on the PejorativITy test partition.
We compare our fine-tuned AlBERTo model (\textit{baseline}) 
against the alternatives
that leverage pejorative word disambiguation.
We evaluate the concatenation and substitution approaches using model$_{pej}$ (\textit{w/ predictions}) and annotators' labels (\textit{w/ gold}) since our corpus contains annotations for pejorative word disambiguation.
The evaluation of our proposed approaches with gold labels defines an upper bound to our pipeline.
We observe a notable improvement over the baseline model for concatenation ($+7$ absolute points) and substitution ($+9$ absolute points) when using model$_{pej}$ predictions.
The improvement significantly increases when both approaches consider gold labels, with a maximum gain of $+19$ absolute points.
These results reflect the effectiveness of our approach and corroborate our initial hypothesis on reducing the false positive rate for misogyny detection.

Table \ref{tab:fp} shows the number of false positives in the three datasets, before and after the inclusion of pejorative information both by concatenation and substitution. 
The decrease of false positives is clear in AMI-2020 and in our PejorativITy test set. In AMI-2018, no decrease is observed.
One of the reasons for this low impact is that
AMI-2018 contains pejorative epithets only in 34 instances out of 1000 (compared to 192 in AMI-2020), therefore we did not expect our approach to have a huge impact on that dataset.

Table~\ref{tab:results} shows the classification performance for misogyny detection on AMI-2018 and AMI-2020.
To assess 
the impact of our pipeline on these corpora, we show the performance of the models both on the test instances that contain 
words in our lexicon (\textbf{epithets}) and on the whole corpora. In particular, we perform fuzzy string matching (Section~\ref{sec:corpus_compilation}) to filter tweets according to this criterion, resulting in 389 (355 train, 34 test)  tweets for AMI-2018 and 605 (413 train, 192 test) tweets for AMI-2020 in the training and test set respectively.
We observe an F$_1$-measure improvement of $+3$ absolute points in AMI-2018 and $+4$ absolute points in AMI-2020 with the concatenation approach.
In contrast, the substitution strategy does not lead to any performance gain.
A possible explanation is the quality of substituted anchors. We provide an example in the next section.
Since AMI corpora mainly contain tweets with explicit misogyny, the limited number of retrieved samples is expected.
For this reason, the observed gain on selected tweets does not impact the overall performance on the original test partition in both corpora (\textbf{whole}).

\begin{table}[t]
\centering
\begin{tabular}{lccc}
\hline
\textbf{Dataset} & \textbf{Baseline} & \textbf{concat.} & \textbf{subst.} \\
\hline
PejorativiITy & \,\,\,25 & \,\,\,16 & \,\,\,21 \\
AMI 2018 & 107 & 107 & 112 \\
AMI 2020 & 127 & 126 & 121 \\
\hline
\end{tabular}
\caption{False positive rates comparison. In the PejorativITy the total number of instance is 96, while in AMI 2018 and 2020 is 1,000.}
\label{tab:fp}
\end{table}

To sum up, our results suggest that the disambiguation of potentially pejorative words is helpful in addressing misogyny detection when targeting ambiguous examples.

\subsection{Qualitative Error Analysis}

We carry out a manual error analysis, by observing misclassified tweets in AMI-2020 epithets and our corpus for the task of misogyny detection. We compare misclassified tweets in the three settings: baseline, concatenation, and substitution.

Regarding the concatenation approach, most of the misclassifications occur when reported misogyny is concerned. The model struggles to recognise when a pejorative epithet is used in a reported speech to condemn a misogynistic attitude and not to address a potential target.
It is worth noticing that if a pejorative connotation is predicted in reported speech, this does not imply that misogyny is predicted.
Consider the following example:

 \begin{myquote}
     \begin{footnotesize}
        \textit{Lei \`e acida perch\'e non ha figli penso che darebbe fastidio a qualsiasi donna. Che schifo.}\footnote{She's peevish because she doesn't have children I think it would bother all women. Disgusting.}
     \end{footnotesize}
 \end{myquote}
 
\begin{table}[t]

    \centering
    \resizebox{1.0\columnwidth}{!}
    {%
    \begin{tabular}{lcccccc}
    \toprule

\bf AMI-2018 &\multicolumn{3}{c}{\textbf{epithets}}&\multicolumn{3}{c}{\textbf{whole}}\\
\textbf{Approach}&\textbf{Macro}&\textbf{Mis.}&\textbf{Not}&\textbf{Macro}&\textbf{Mis.}&\textbf{Not}\\
\midrule

baseline  &0.79&0.77&0.81&0.86&0.87&0.85\\
concatenation &\bf0.82&0.81&0.83&\bf 0.86&0.88&0.85\\

substitution &0.79&0.79&0.80&0.86&0.87&0.84\\
         \midrule
\bf AMI-2020&\multicolumn{3}{c}{\textbf{epithets}}&\multicolumn{3}{c}{\textbf{
  whole}}\\
\textbf{Approach}&\textbf{Macro}&\textbf{Mis.}&\textbf{Not}&\textbf{Macro}&\textbf{Mis.}&\textbf{Not}\\
\midrule

baseline  &0.77&0.74&0.81&0.82&0.84&0.81\\
concatenation &\bf0.81&0.77&0.84&\bf0.83&0.84&0.82\\
substitution &0.77&0.73&0.81&0.82&0.84&0.81\\
         \bottomrule
    \end{tabular}
    }
    \caption{Macro and per-class F$_1$-measure on AMI-2018 and AMI-2020 concerning misogyny detection. We report metrics for each corpus (\textbf{whole}) and their subset containing words in our lexicon (\textbf{epithets}).}
    \label{tab:results}
\end{table}

In this example, the author of the tweet criticises a reported misogynous sentence. Even if \textit{acida} is correctly predicted as pejorative, the model still gets the correct prediction that the sentence is non-misogynous.
Another observed pattern of misclassification is when the target of the pejorative epithet is a man. In this case, the tweet should not be considered misogynous, although it contains a pejorative word from our lexicon. This bias is introduced due to the annotation of pejorative epithets against men as pejorative.
Overall, the overlap between tweets classified as containing pejorative words and those classified as misogynous is of 26 tweets in the PejorativITy test set (out of 96), 12 tweets out of 34 in the AMI2018\_epithets, and 67 out of 192 in AMI2020\_epithets. We highlight this aspect to show that model\_$mis$ does not necessarily learn to classify misogyny according to model\_$pej$'s outcome.

Regarding the substitution approach, we observe that a wrong pejorative prediction of lexicon words affects the prediction of misogyny.
The following example:

 \begin{myquote}
     \begin{footnotesize}
        \textit{Ma la balena con gli shorts cortissimi invece \`e vittima del patriarkato e pu\`o vestirsi come vuole?}\footnote{That whale/fat girl with very short pants is a victim of the patriarchy and can dress up as she wants?}
     \end{footnotesize}
 \end{myquote}

is correctly classified by the 
baseline model. 
A misclassification of 
the word \textit{balena},which model$_{pej}$ predicts 
as neutral, %
causes confusion in both enriched models.

\begin{table*}[t]
    \centering
\resizebox{1.0\linewidth}{!}{
\begin{tabular}{llrrrr}
\toprule
                                           & \multicolumn{1}{l}{}           & \multicolumn{2}{c}{\bf pretrained}                & \multicolumn{2}{c}{\bf Fine-tuned}                 \\
\multicolumn{1}{c}{\bf Lexicon}                 & \bf Anchor                         & \multicolumn{1}{c}\ {Pejorative} & \multicolumn{1}{c} {Neutral} & \multicolumn{1}{c}{Pejorative} & \multicolumn{1}{c}{Neutral} \\ \midrule
 & \np{aspra}                        &  \ares{0.27}{0.12}                     &  \ares{0.27}{0.14}                      &    \ares{0.09}{0.12}                   &    \ares{0.29}{0.10}                    \\
acida                       & \pos{intrattabile}                   & \ares{0.28}{0.12}                     &  \ares{0.28}{0.14}                     &     \ares{0.28}{0.05}                  &      \ares{0.27}{0.07}                  \\
\multicolumn{1}{c}{}                       & \pos{stronza}                        & \ares{0.31}{0.14}                     & \ares{0.31}{0.17}                      &  \ares{0.53}{0.12}                     &    \ares{0.23}{0.15}                    \\ \midrule
\multirow{3}{*}{balena}                    & \np{balenare}                       &    \ares{0.26}{0.12}                   &   \ares{0.30}{0.10}                     &    \ares{0.19}{0.10}                   &    \ares{0.44}{0.08}                    \\
                                           & \np{cetaceo}                        &    \ares{0.22}{0.12}                   &   \ares{0.26}{0.09}                     &   \ares{0.04}{0.10}                    &     \ares{0.36}{0.10}                   \\
                                           & \pos{grassa}                         &  \ares{0.19}{0.12}                     &   \ares{0.22}{0.09}                     &   \ares{0.29}{0.09}                    &    \ares{0.07}{0.07}                    \\ \midrule
\multirow{3}{*}{cagna}                     & \np{canide}                         &   \ares{0.43}{0.15}                    &   \ares{0.29}{0.15}                     &   \ares{0.08}{0.05}                    &     \ares{0.25}{0.06}                   \\
                                           & \pos{donna di facili costumi}        &  \ares{0.42}{0.13}                     &   \ares{0.27}{0.15}                     &    \ares{0.30}{0.04}                   &   \ares{0.21}{0.09}                     \\
                                           & \pos{troia}                          &  \ares{0.41}{0.16}                     &   \ares{0.26}{0.16}                     &   \ares{0.57}{0.08}                    &     \ares{0.21}{0.10}                   \\ \midrule
\multirow{4}{*}{cesso}                     & \np{water}                          &  \ares{0.37}{0.14}                     &   \ares{0.37}{0.13}                     &    \ares{0.08}{0.06}                   &    \ares{0.26}{0.08}                   \\
                                           & \np{bagno  }                        &    \ares{0.39}{0.14}                   &   \ares{0.41}{0.13}                     &   \ares{0.07}{0.06}                    &        \ares{0.35}{0.10}                \\
                                           & \np{toilette}                       &  \ares{0.37}{0.13}                     &  \ares{0.39}{0.12}                      &   \ares{0.09}{0.05}                    &     \ares{0.30}{0.08}                   \\
                                           & \pos{brutta }                       &   \ares{0.39}{0.15}                    &  \ares{0.40}{0.13}                      &   \ares{0.43}{0.07}                    &     \ares{0.16}{0.09}                   \\ \midrule
\multirow{3}{*}{lesbica}                   
                                           & \np{donna a cui piacciono le donne} &  \ares{0.40}{0.13}                     &   \ares{0.42}{0.16}                     &     \ares{0.28}{0.05}                  &   \ares{0.34}{0.09}                     \\
                                           & \pos{schifosa}                       &  \ares{0.32}{0.15}                     &  \ares{0.32}{0.17}                      &    \ares{0.30}{0.09}                   &    \ares{0.18}{0.06}                    \\ \midrule
\multirow{3}{*}{vacca}                     & \np{bovino}                        &   \ares{0.31}{0.14}                    &   \ares{0.25}{0.12}                     &   \ares{0.10}{0.07}                    &      \ares{0.22}{0.07}                  \\
                                           & \pos{donna di facili costumi}        &   \ares{0.35}{0.12}                    &  \ares{0.29}{0.12}                      &   \ares{0.27}{0.05}                    &       \ares{0.20}{0.08}                 \\
                                           & \pos{troia}                        &  \ares{0.35}{0.14}                     &  \ares{0.29}{0.13}                      &     \ares{0.50}{0.09}                  &     \ares{0.25}{0.14}                   \\ \bottomrule
\end{tabular}%
}
    \caption{Average cosine similarity between lexicon word embeddings and both \pos{pejorative} and \np{neutral} anchor word embeddings in pejorative and neutral samples. Embeddings extracted from both the pretrained and the fine-tuned AlBERTo model.}
    \label{tab:cosine_sim}
\end{table*}

\section{Analysis of Contextualised Word Embeddings}

To investigate the semantic knowledge of the AlBERTo pretrained language model~\cite{PolignanoEtAlCLIC2019} about the pejorative epithets and to evaluate how fine-tuning affects its knowledge (\textbf{RQ3}), we extract and analyse the contextualised word embeddings of our lexicon words. 

To extract these embeddings, we perform fuzzy string matching on input tweets to retrieve the tokenized text span corresponding to lexicon words.
We use fuzzy string matching to address all representations of a lexicon word (e.g., \textit{balena} and \textit{balenare}).
It is worth noticing that the retrieved text span may contain multiple tokens according to the employed tokenization process. 
In our scenario, the AlBERTo model employs the sentencepiece tokenizer~\cite{kudo-richardson-2018-sentencepiece}, the common tokenization process for transformer models.
For instance, the lexicon word \textit{balena} is tokenized to the
[\texttt{balen}, \texttt{\#\#a}] text span.
We then use these text spans to aggregate the corresponding word embeddings. 
We define the word embedding of a lexicon word as the average of the AlBERTo token embeddings in the retrieved text span. Considering \textit{balena}, we define its word embedding by extracting the embeddings of \texttt{balen} and \texttt{\#\#a} and computing their average.

We compute the average cosine similarity between lexicon words and their corresponding neutral and pejorative anchors. 
To carry out our analysis, we consider lexicon words from PejorativITy with several neutral and offensive anchors: \textit{acida}, \textit{balena}, \textit{cagna}, \textit{cesso}, \textit{lesbica}, and \textit{vacca}.

Table~\ref{tab:cosine_sim} reports the results on PejorativITy comparing the pretrained AlBERTo model and its fine-tuned version. The pretrained model does not discriminate between \np{neutral} and \pos{offensive} anchors in pejorative and neutral samples. For instance, the average cosine similarity between \textit{acida} and its pejorative anchor \pos{stronza} is 0.31 in both class samples. 
In contrast, our fine-tuned AlBERTo model shows relevant discrepancies when considering lexicon word embeddings in pejorative and neutral samples. 
For instance, the similarity between \textit{acida} and its neutral anchor \np{aspra} is 0.09 in pejorative samples and 0.29 in neutral ones.
In contrast, the similarity between \textit{acida} and its pejorative anchor \pos{stronza} is significantly higher in pejorative samples (0.53), compared to neutral ones (0.23). 
This is an indicator that our fine-tuned model acquired knowledge about \textit{acida} being a synonym of \pos{stronza} when used in a pejorative way.
We observe similar trends in all other selected lexicon words.


The average similarity with \pos{pejorative} anchors in pejorative and neutral samples using pretrained embeddings is 0.34 and 0.29, respectively. 
In contrast, the similarities are 0.39 and 0.20 using fine-tuned embeddings.
These results show \pos{pejorative} anchors have higher similarity with lexicon words in pejorative samples.
Likewise, \np{neutral} anchors with lexicon words in neutral samples: the average similarity in pejorative and neutral samples using pretrained embedding is 0.33 and 0.34, respectively, while the similarities are 0.11 and 0.31 using fine-tuned embeddings.


To prevent a biased analysis where our fine-tuned model learns a representation of anchors similar to that of lexicon words, we compute each anchor frequency in our corpus. 
The computed frequency for each anchor is close to zero, thus excluding potential biases in our analysis.



\begin{table*}[t]

\centering
\small
\resizebox{1.0\linewidth}{!}{%
\begin{tabular}{r@{\hspace{1mm}}p{0.27\linewidth}p{0.27\linewidth}p{0.4\linewidth}}
\toprule
& \bf Tweet & \bf Translation & \bf Generated Disambiguation \\
\midrule

1. & \textit{Tra il non normalizzare e il dargli della \underline{balena} ci sta l’oceano pacifico.}  & \textit{Between the non-normalisation and calling her \underline{whale} there is the Pacific Ocean.} & \textit{The meaning of the word \underline{balena} in this sentence is to indicate something large, imposing, and difficult to overcome or face, similar to the Pacific Ocean.} \\[0.1cm]

2. & \textit{Non voglio una \underline{cagna} un cane ce l'ho giaaaa} & I don't want a \underline{female dog/bitch}, I have a dog already. & The meaning of the word 
\underline{cagna} in this sentence is “ragazza” (girl). The word \underline{cagna} is a slang term used to refer to a girl. In this context, the phrase means "I don't want a girl, I already have a dog." \\[0.1cm]

3. & \textit{Chi sei tu per dare \underline{cagna} e zoccola alle persone?} & \textit{Who are you to address people as \underline{bitch} and whore?} & \textit{In this case, the author of the sentence is calling the other user a \underline{bitch} probably to express anger or contempt toward her.} \\[0.1cm]

4. & \textit{Madonna mia che \underline{cozza} questa!!} & \textit{What an \underline{ugly girl}!} & The meaning of the word \underline{cozza} in this phrase is to express surprise or amazement. Specifically, the phrase can be literally translated as "My God, how beautiful this is!".\\ 

\bottomrule
\end{tabular}%
}
\caption{Examples of Mistral pejorative word disambiguation. \underline{Lexicon words} are underlined.}
\label{tab:mistral}
\end{table*}

\section{Analysis of Large Language Models}

We prompt open-source generative large language models (LLMs) to investigate their knowledge of pejorative words. 
Specifically, we aim to find out whether they are able to clarify the meaning of our lexicon words (\textbf{RQ3}).

We follow a zero-shot prompting approach without any training data for the downstream task~\cite{Liu:2023}. 
We design the following prompt that asks the meaning of a pejorative \emph{\{word\}} in a given \emph{\{sentence\}}. 
We prompt the LLMs for each test instance in our corpus. The prompt is given as follows:

\begin{myquote}
 \begin{footnotesize}
    \texttt{
    [INST] Di seguito \`e riportata un'istruzione che descrive un task. Scrivete una risposta che completi adeguatamente la richiesta.
    \\
    \#\#\# Istruzione:
    \\
    Qual \`e il significato della parola ''\{word\}'' in questa frase?
    \\
    ''\{sentence\}''[/INST]
    \\
    \#\#\# Risposta:
    }
 \end{footnotesize}
\end{myquote}
 
The translation in English would be:

\begin{myquote}
     \begin{footnotesize}
     \texttt{
        [INST] Below there is an instruction describing a task. Write a response that completes the request appropriately.
        \\
        \#\#\# Instruction:
        \\
        What is the meaning of the word "\{word\}" in this sentence?
        \\
        ''\{sentence\}''[/INST]
        \\
        \#\#\# Response:
    }
     \end{footnotesize}
\end{myquote}
 
We use three open-source LLMs for our analysis: 

\paragraph{LlaMa:} LlaMa is a decoder-based language model pretrained on publicly available data collections \cite{Llama:2023}. Since LlaMa does not support Italian, we employ Camoscio\footnote{https://github.com/teelinsan/camoscio}, an Italian instruction-tuned LlaMa model.

\paragraph{LlaMa2:} LlaMa2\footnote{https://huggingface.co/meta-llama/Llama-2-7b-hf} is an optimized version of LlaMa by increasing context length from 2048 to 4096, and applying group-query attention~\cite{Llama2:2023}. The majority of the training data of LlaMa2 is in English, but it still responds to Italian prompts due to a small amount of training data in Italian.

\paragraph{Mistral:} MistralAI\footnote{https://huggingface.co/mistralai} is based on LlaMa2 but exhibits superior performance due to the employed attention mechanisms such as group-query attention and sliding-window attention~\cite{Mistral:2023}. The details of the corpora used in training are not given, yet it responds to Italian prompts.

For all models, we select the \textit{7b} model version with 8-bit weights due to hardware constraints.
We apply Beam Search for text generation with the following hyperparameters. 
The temperature is set to 0.2, the number of beams is set to 4 with a top-p value of 0.75, the output length is set to 300 tokens with a repetition penalty of 1.8. 
The analysis is conducted on 4 GPUs (NVIDIA GeForce RTX 2080 Ti).

\subsection{Qualitative Analysis}
We manually inspect all the responses generated by the three LLMs.
\paragraph{Mistral.}
We observe that Mistral is the best in disambiguating the connotations of lexicon words, especially when it comes to neutral senses.
For instance, it correctly disambiguates when \textit{balena} refers to the animal and when to the verb \textit{balenare}. 

However, Mistral struggles when the term \textit{balena} is used pejoratively.
Consider Example~1 in Table~\ref{tab:mistral}.
Mistral gets the idea that \textit{balena} is used as a metaphor for something big, but it does not link its meaning to being overweight.

Mistral is remarkably good at capturing irony as well.
Consider Example~2 in Table~\ref{tab:mistral}.
While this example caused trouble to human annotators for the lack of context, Mistral is confident in identifying the pejorative connotation of the lexicon word \textit{cagna}.


Although performing very well, Mistral struggles with reported speech, too. 
Consider Example~3 in Table~\ref{tab:mistral}.
While Mistral correctly identifies the pejorative connotation, 
it fails to   
understand that the author of the tweet is condemning, not enforcing, a misogynistic statement.


Moreover, in some cases, Mistral makes up meanings.
For instance, Mistral defines \textit{cavalla} (horse / ugly and tall woman) as a \textit{``a painful surprise''}, while it defines \textit{cozza} (mussel / ugly, clingy) as \textit{``impatiently waiting''}.\footnote{In the sentence "Sta cozza non vedeva l'ora", translated as "That ugly girl couldn't wait"}
A possible explanation is that Mistral uses the semantics of the whole sentence to generate a definition of lexicon words.
In some other cases, Mistral generates the opposite meaning.
In Example~4 of  Table~\ref{tab:mistral}, Mistral defines \textit{cozza} as \textit{``surprisingly beautiful''}.



\paragraph{Llama and Camoscio.} Neither model shows an adequate performance in disambiguating lexicon words. 
In most cases, both models produce the following answer: \textit{``the word \{word\} means \{word\}''}, which is not useful for disambiguation.
\medskip

Our analysis suggests that off-the-shelf instruction-tuned LLMs have ample room for improvement concerning pejorative word disambiguation.
A fine-tuning phase on the task could address the highlighted issues.
However, we believe that a detailed analysis of instruction-tuned LLMs on our proposed pipeline deserves a separate study. We leave this analysis as future work.





\section{Conclusions}

We introduce pejorative word disambiguation as a preliminary step for misogyny detection to reduce the error rate of classification models on polysemic words that can serve as pejorative epithets.
For this purpose, we build a lexicon of polysemic words with both pejorative and neutral connotations and use it to compile a novel corpus of 1,200 manually expert-annotated Italian tweets for pejorative word disambiguation and misogyny detection.
We validate our pipeline by evaluating AlBERTo~\cite{PolignanoEtAlCLIC2019} on our corpus and on two benchmark corpora in Italian: AMI-2018 \cite{fersininozzarosso-overview-2018} and AMI-2020 \cite{Fersini2020}. 
We explore two approaches to inject pejorativity information: concatenation and substitution.
Our results show that the disambiguation of potentially pejorative words leads to notable classification improvements in all testing scenarios.
Furthermore, we analyse the word embedding representation of AlBERTo and show that the encoding of lexicon words is closer to their ground-truth connotation after fine-tuning.
Lastly, we qualitatively analyse several off-the-shelf instruction-tuned LLMs on pejorative word disambiguation to evaluate their capabilities, showing that there is ample room for improvement.

Future research directions include the extension of our pipeline to automatically extract potentially pejorative words at the span level;
the application of knowledge bases like ConceptNet~\cite{conceptnet} for pejorative word disambiguation; and 
the implementation of instruction-tuned LLMs in our pipeline. Moreover, we plan to expand this work towards other languages, and the cultures behind them. 
Our aim is to carry out a cross-cultural analysis on the differences in terms of pejorative terms for misogyny across 
cultures with different perspectives towards women rights and feminism.

\section{Ethical Considerations}
We use publicly available tweets to collect our corpus.
All data collection adheres to Twitter's terms of service and privacy policies. 
As this research involves the analysis of publicly available tweets, we do not seek explicit consent from individual users. 
Nevertheless, we make every effort to protect the anonymity of all individuals mentioned or quoted in this work: Any reported example is carefully selected to avoid identifying specific users or victims.

\section{Limitations}

\paragraph{Language.}

In our study, we only focus on the Italian language.
While this choice does not limit the applicability of our contributions, we are aware that including other languages could strengthen the impact of our results. We leave this extension as future work.

\paragraph{Corpus}
Although our lexicon covers a wide variety of words that can serve as pejorative epithets for women, it is not an exhaustive list, as we have discarded all the terms that are not polysemic and that are used only with one connotation (either positively or negatively) on Twitter.

Only 100 tweets are annotated by six annotators, while the remaining 1,100 are labelled by only one annotator. Although we select an expert with an interdisciplinary background in linguistics, gender studies and NLP to carry out all the annotations, their personal biases, opinions, or interpretations can lead to skewed or one-sided data.


\paragraph{Approaches.}

A limitation of our study concerns the substitution approach.
First of all, some words have more than one neutral anchor words. This is the case of \textit{balena}, which has two neutral anchors: \textit{balenare} (to flash) and \textit{cetaceo} (sea mammal). 
In neutral examples, we substitute \textit{balena} with both anchors. This process may alter the semantic meaning of the tweet since only one anchor is suitable for substitution.
Moreover, in some cases, we replace a lexicon word with anchors that do not have the same meaning. 
For instance, the neutral anchor of \textit{acida} is \textit{aspra} (\textit{sour}).
However, expressions like \textit{sour beer} or \textit{sour cream} do not have a valid anchor replacement.
Therefore, replacing \textit{aspra} with \textit{acida} is not an appropriate substitution.


\paragraph{Models.}
We only employ AlBERTo to carry out our experiments.
However, several other models might lead to different results. Therefore, our experiments are not sufficient to generalise the results of our analysis to all encoder-based models.

\paragraph{Prompting.}
The most popular generative models ---GPT family--- have not been included in this study, although they could have shown promising capacities in disambiguating the senses of our polysemic words. Nevertheless, we intentionally exclude them from this study, as we want to focus on open-source models only.

\section*{Acknowledgements}
A. Muti’s research is carried out under project “DL4AMI–Deep Learning models for Automatic Misogyny Identification”, in the framework of Progetti di formazione per la
ricerca: Big Data per una regione europea più ecologica, `digitale e resiliente—Alma Mater Studiorum–Universita di `Bologna, Ref. 2021-15854.
The work of F. Ruggeri is supported by the European Union's Horizon Europe research and innovation programme under GA 101070000.
We thank to Umitcan Sahin for his support during corpus compilation.

\nocite{*}
\section{Bibliographical References}\label{sec:reference}

\bibliographystyle{lrec-coling2024-natbib}
\bibliography{bibliography}

\section{Language Resource References}
\label{lr:ref}
\bibliographystylelanguageresource{lrec-coling2024-natbib}
\bibliographylanguageresource{languageresource}

\end{document}